\pdfoutput=1
\documentclass{article} 
\usepackage{glen_article_style,times}

\usepackage[printonlyused,nohyperlinks]{acronym}
\usepackage{subcaption}

\usepackage{amssymb}
\usepackage{amsmath}
\usepackage{xspace}
\usepackage{hyperref}
\usepackage{booktabs}
\usepackage{float}

\usepackage{graphicx}
\usepackage{algpseudocode}
\usepackage{algorithm}
\usepackage{multirow}
\usepackage{verbatim}
\usepackage{array}
\usepackage{tikz}
\usetikzlibrary{calc}
\usetikzlibrary{decorations.pathmorphing}
\usepackage{caption}
\usepackage{siunitx}
\usepackage{listings}
\usepackage{color}
 
\definecolor{codegreen}{rgb}{0,0.6,0}
\definecolor{codegray}{rgb}{0.5,0.5,0.5}
\definecolor{codepurple}{rgb}{0.58,0,0.82}
\definecolor{backcolour}{rgb}{0.95,0.95,0.92}
 
\lstdefinestyle{mystyle}{
    backgroundcolor=\color{backcolour},   
    commentstyle=\color{codegreen},
    keywordstyle=\color{magenta},
    numberstyle=\tiny\color{codegray},
    stringstyle=\color{codepurple},
    basicstyle=\footnotesize,
    breakatwhitespace=false,         
    breaklines=true,                 
    captionpos=b,                    
    keepspaces=true,                 
    numbers=left,                    
    numbersep=5pt,                  
    showspaces=false,                
    showstringspaces=false,
    showtabs=false,                  
    tabsize=2
}
 
\title{Terrain RL Simulator}

\author{
Glen Berseth$^1$ 
\And 
Xue Bin Peng$^2$ 
\And 
Michiel van de Panne$^1$ \\
\And
$^1$University of British Columbia, $^2$UC Berkeley \\
\{gberseth,van\}@cs.ubc.ca
}

\newcommand{\todo}[1]{}
\newcommand{\glen}[1]{}

\newcommand{\refSection}[1]{Section: \ref{#1}}
\newcommand{\refFigure}[1]{Figure~\ref{#1}}

\newcommand{\valueWithUnits}[2]{#1~\scriptsize #2\normalsize}

\newcommand{\bipedController}[0]{\emph{biped}\xspace}
\newcommand{\dogController}[0]{\emph{dog}\xspace}
\newcommand{\raptorController}[0]{\emph{dog}\xspace}

\newcommand{\deepRL}{\ac{DRL}\xspace}
\newcommand{\HRL}{\ac{HRL}\xspace}

\newcommand{\RL}{\ac{RL}\xspace}

\newcommand{\progRL}[0]{\ac{PLAiD}\xspace}

\newcommand{\deepLoco}{\ac{DeepLoco}\xspace}
\newcommand{\terrainRL}{\ac{terrainRL}\xspace}
\newcommand{\SIMBICON}{\ac{SIMBICON}\xspace}
\newcommand{\FSM}{\ac{FSM}\xspace}
\newcommand{\environments}{\emph{environments}\xspace}
\newcommand{\terrainRLSim}{\emph{TerrainRLSim}\xspace}

\newcommand{\agent}{agent\xspace}

\newcommand{\Flat}{\emph{flat}\xspace}
\newcommand{\incline}{\emph{incline}\xspace}

\newcommand{\gaps}{\emph{gaps}\xspace}

\newcommand{\slopes}{\emph{slopes}\xspace}
\newcommand{\steps}{\emph{steps}\xspace}
\newcommand{\mixed}{\emph{mixed}\xspace}
\newcommand{\forest}{\emph{forest}\xspace}
\newcommand{\pathFollowing}{\emph{follow-path}\xspace}
\newcommand{\dynamicObstacles}{\emph{dynamic obstacles}\xspace}
\newcommand{\largeBlocks}{\emph{large-blocks}\xspace}
\newcommand{\soccer}{\emph{soccer}\xspace}
\newcommand{\slopesMixed}{\texttt{slopes-mixed}}
\newcommand{\narrowGaps}{\texttt{narrow-gaps}}
\newcommand{\tightGaps}{\texttt{tight-gaps}}
\newcommand{\slopesGaps}{\texttt{slopes-gaps}}
\newcommand{\slopesWalls}{\texttt{slopes-walls}}
\newcommand{\slopesSteps}{\texttt{slopes-steps}}
\newcommand{\variableSteps}{\texttt{variable-steps}}
\newcommand{\numEnvironments}{89}

\begin{document}

\maketitle

\acrodef{UI}{user interface}
\acrodef{UBC}{University of British Columbia}
\acrodef{MDP}{Markov Dynamic Processes}


\acrodef{ANOVA}[ANOVA]{Analysis of Variance\acroextra{, a set of
  statistical techniques to identify sources of variability between groups}}
\acrodef{API}{application programming interface}
\acrodef{CTAN}{\acroextra{The }Common \TeX\ Archive Network}
\acrodef{DOI}{Document Object Identifier\acroextra{ (see
    \url{http://doi.org})}}
\acrodef{GPS}[GPS]{Graduate and Postdoctoral Studies}
\acrodef{PDF}{Portable Document Format}
\acrodef{RCS}[RCS]{Revision control system\acroextra{, a software
    tool for tracking changes to a set of files}}
\acrodef{TLX}[TLX]{Task Load Index\acroextra{, an instrument for gauging
  the subjective mental workload experienced by a human in performing
  a task}}
\acrodef{UML}{Unified Modelling Language\acroextra{, a visual language
    for modelling the structure of software artefacts}}
\acrodef{URL}{Unique Resource Locator\acroextra{, used to describe a
    means for obtaining some resource on the world wide web}}
\acrodef{W3C}[W3C]{\acroextra{the }World Wide Web Consortium\acroextra{,
    the standards body for web technologies}}
\acrodef{XML}{Extensible Markup Language}
\acrodef{MBAE}{Model-Based Action Exploration}
\acrodef{SMBAE}{Stochastic Model-Based Action Exploration}
\acrodef{DRL}{Deep Reinforcement Learning}
\acrodef{HRL}{Hierarchical Reinforcement Learning}
\acrodef{DDPG}{Deep Deterministic Policy Gradient}
\acrodef{CACLA}{Continuous Actor Critic Learning Automaton}
\acrodef{HLC}{High-Level Controller}
\acrodef{LLC}{Low-Level Controller}
\acrodef{ReLU}{Rectified Linear Unit}
\acrodef{PPO}{Proximal Policy Optimization}
\acrodef{EPG}{Expected Policy Gradient}
\acrodef{DPG}{Deterministic Policy Gradient}
\acrodef{DYNA}{DYNA}
\acrodef{GAE}{Generalized Advantage Estimation}
\acrodef{RL}{Rienforcement Learning}
\acrodef{SVG}{Stochastic Value Gradients}
\acrodef{GAN}{Generative Advasarial Network}
\acrodef{cGAN}{Conditional Generative Advasarial Network}
\acrodef{MSE}{Mean Squared Error}
\acrodef{DeepLoco}{Deep Locomotion}
\acrodef{terrainRL}{Terrain Adaptive Locomotion}
\acrodef{SIMBICON}{Simple Biped Controller}
\acrodef{FSM}{Finite State Machine}
\acrodef{RBF}{Radial Basis Function}
\acrodef{PLAiD}{Progressive Learning and Integration via Distillation}

%

\begin{abstract}

We provide $\numEnvironments$ challenging simulation environments that range in difficulty.
The difficulty of solving a task is linked not only to the number of dimensions in the action space but also to the size and shape of the distribution of configurations the \agent experiences.
Therefore, we are releasing a number of simulation \environments that include randomly generated terrain.
The library also provides simple mechanisms to create new environments with different \agent morphologies and the option to modify the distribution of generated terrain.	
We believe using these and other more complex simulations will help push the field closer to creating human-level intelligence.

\end{abstract}

\section{Introduction}
\label{sec:Intro}


Research in \deepRL has grown significantly in recent years, and so to has the demand for simulated environments that can be used to evaluate \deepRL algorithms.
Some environments serve as standard benchmarks to evaluate the performance of \deepRL algorithms by ensuring the simulation and reward function are the same across papers.
More custom environments have also been created as challenges for researchers and practitioners to achieve higher quality results.
Although, many \environments have been created, not enough is truly known about the difficulty of the \environments.
Many aspects of control problems make them challenging to solve: sparse/delayed rewards, large number of dimensions in the control space, complex dynamics, etc. 
For example, getting a simulated biped to walk and be robust to perturbations could be challenging, however, simple control structures were created to facilitate this problem years ago~\citep{Yin07,1241826,770006,973365}.
The \environments included in openAIGym have similar and simpler control problems that have recently been solved using methods less complicated than \deepRL.
These methods include using \ac{RBF}~\citep{DBLP:journals/corr/RajeswaranLTK17}\footnote{Simple Nearest Neighbor Policy Method for Continuous Control Tasks} and random search in the network parameter space~\citep{2017arXiv170303864S,2018arXiv180307055M}.
These papers note that the improvements in \deepRL methods in the recent years could be focusing on the challenges related to optimization, not exploration and discovery of good actions. 
Although, this might be possible the authors view the prospects of finding solutions to these problems using less complex methods a sign that the environments used are too simple.

In \deepRL we not only want to push the boundaries of how efficiently we can solve problems but to also make strides in solving new challenging tasks that benefit from task generalization.
What makes a problem challenging is not only related to the control capabilities but also the affordances available in the environment~\citep{gibson1979ecological}.
Therefore, we need to shape the affordances available to the agent as well to affect the difficulty of a task.
We provide TerrainRLSIM\footnote{\url{https://github.com/UBCMOCCA/TerrainRLSim}}, a library with many difficult control tasks, many that have not been seen before.
We provide an easy mechanism to create even more challenging environments, via parameterized terrain generation, and encourage people to create more challenging environments.



\section{Related Work}
\label{sec:related_work}

There are a number of similar libraries for evaluating reinforcement learning methods.
The Arcade Learning Environment is one of the first sets of \environments that was used to show the effectiveness of \deepRL on tasks with high dimensional observations~\citep{DBLP:journals/corr/abs-1207-4708}. 
The OpenAIGym contains a collection of discrete action as well as continuous action tasks~\citep{DBLP:journals/corr/BrockmanCPSSTZ16}. 
OpenAI Roboschool is a version of OpenAIGym where a number of the \environments have been recreated using Bullet instead of Mujoco~\footnote{https://github.com/openai/roboschool}.
DeepMind recently released a new character motion control library (DeepMind Control Suite) that includes control problems similar to openAIGym with additional \environments for mocap imitation~\citep{2018arXiv180100690T}. 
The OpenAI Universe is a different, large set of \environments created with the goal of it being used to create a general agent that can play a large number of games competitively~\footnote{https://blog.openai.com/universe/}. 
The DeepMind-lab is another set of \environments that focuses on using visual inputs as observations, the visual input provides the agent with partial information of the environment state~\citep{DBLP:journals/corr/BeattieLTWWKLGV16}. 
Expanding upon the partially observable \environments is ELF that includes a novel RTS game.~\citep{NIPS2017_6859} 
	
We provide a set of \environments that include tasks similar to openAIGym and the DeepMind Control Suite and new more challenging control problems.
The simulation \environments use Bullet~\citep{Bullet} an open source free simulator where many continuous control libraries use Mujoco~\citep{6386109} a 
non-free, closed source piece of software.
Many \environments include terrain features in the observation. 
In the \environments with terrain state features the \agent navigates over terrain that is randomly generated between episodes. 
As a result not only does the \agent need to learn to locomote but it also needs to learn how to perceive its environment and avoid obstacles and rough terrain.
Some \environments have been so challenging they could only be solved with \HRL techniques.
We provide these in hopes more will continue work in the area of \HRL.
Additional extra difficult \environments are included that have never been solved.
Many of these difficult tasks were created while working on other projects but we were not able to produce controllers to solve these problems, or the controllers produced were not of sufficient quality.
Last, there are different actuation models to choose from.
Most libraries only offer torques as a means to actuate and control the \agent's movement.
We include a variety of control options such as torques, desired velocities, desired position and muscle-based control.

\section{TerrainRLSim}
\label{sec:framework}

\begin{figure}[h!]
\centering
\begin{minipage}{.35\linewidth}
\subcaptionbox{Character state features}{   \includegraphics[width=0.96\columnwidth]{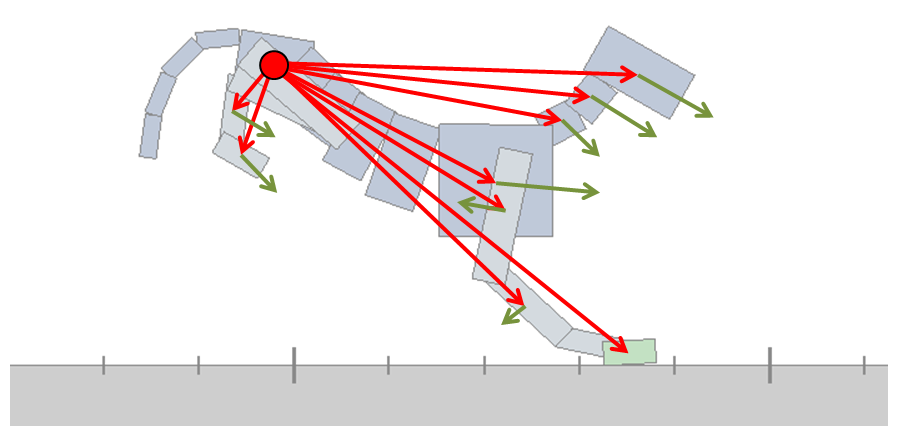}}
\end{minipage}
\begin{minipage}{.55\linewidth}
\subcaptionbox{Terrain state features}{   \includegraphics[width=0.96\columnwidth]{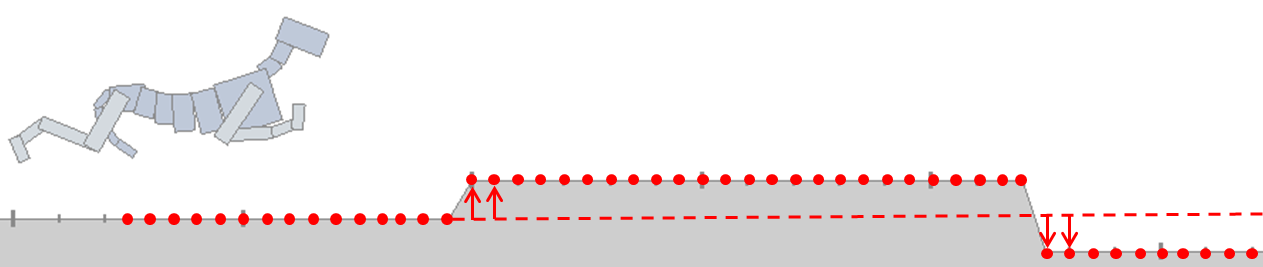}}\\
\\
\subcaptionbox{All state features}{   \includegraphics[width=0.96\columnwidth]{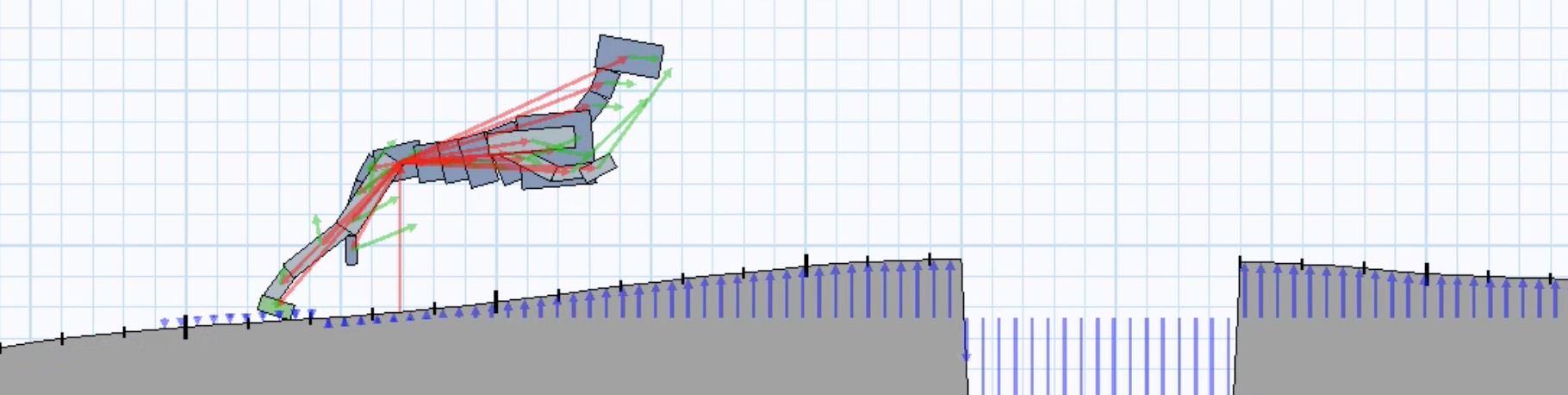}}\\	 
\end{minipage}
\caption{\agent state features from simulation.}
\label{fig:state-features}
\end{figure}

The API closely follows the openAI Gym interface.
We include a mechanism to set the random seed for the simulation, allowing for reproducible simulations.
Many \environments include state features for the local terrain around an egocentric \agent.
The observation produced by the simulation always puts these terrain features first, for example, $(<terrain-features> || <\agent-features>)$, all as a single vector.
The observation can be sliced into multiple parts, allowing only the terrain features to be passed through convolution layers.
The software uses the Bullet Physics library~\citep{Bullet} an open source physics simulator.
The simulation performance depends primarily on the efficiency of Bullet which is highly optimized. Overall, the simulation is fast and supports different kinds of action parameterizations (torque, velocity, pd and muscle activations).
The state features used for an \agent are visualized in~\ref{fig:state-features}.

\todo{Details about simulation, 3000hz, character files, terrain files. How the terrain is generated}

We show a simplified code snippet of how to load and simulated a number of trajectories/epochs of the simulation.
\begin{lstlisting}[language=Python, caption=Example code for loading and simulating an environment.]

import terrainRLSim
import numpy as np
import matplotlib.pyplot as plt
import json

if __name__ == '__main__':

    envs_list = terrainRLSim.getEnvsList()
    print ("# of envs: ", len(envs_list))
    print ("Envs:\n", json.dumps(envs_list, sort_keys=True, indent=4))
    env = terrainRLSim.getEnv(env_name="PD_Biped3D_FULL_Imitate-Steps-v0", render=True)
    
    env.reset()
    actionSpace = env.getActionSpace()
    env.setRandomSeed(1234)
    
    actions = []
    action = ((actionSpace.getMaximum() - actionSpace.getMinimum()) * np.random.uniform(size=actionSpace.getMinimum().shape[0])  ) + actionSpace.getMinimum()
    actions.append(action)            
    
    print("observation_space: ", env.observation_space.getMaximum())
    
    for e in range(10):
        env.reset()
        
        for t in range(100):
            observation, reward,  done, info = env.step(actions)
            print ("Done: ", done)
            if ( done ):
                break
           
            #### LLC states. If there is an LLC
            # llc_state = env.getLLCState()
            # print ("LLC state:", llc_state.shape)
            
            ## Get and vis terrain data
            """
                img_ = np.reshape(states[0][:1024], (32,32))
                print("img_ shape", img_.shape)
                plt.imshow(img_)
                plt.show()
            """
            
            print ("Agent state: ", state)
        
    env.finish()
    print (env)
\end{lstlisting}

The terrain in the environment is randomly generated and controlled via a set of parameters that are defined in a \textit{terrain} file. For example, the terrain file contains parameters that determine the random distance between gaps, the depth of the gap and the width of the gap. 
There are similar settings for many types of obstacles.
Here we give an example of a terrain file.

\begin{lstlisting}[language=Python, caption=Example terrain parameter file.]

{
"Type": "var2d_slopes_mixed",

"Params": [
	{
		"GapSpacingMin": 3,
		"GapSpacingMax": 4,
		"GapWMin": 0.3,
		"GapWMax": 0.5,
		"GapHMin": -2,
		"GapHMax": -2,

		"WallSpacingMin": 3,
		"WallSpacingMax": 4,
		"WallWMin": 0.2,
		"WallWMax": 0.2,
		"WallHMin": 0.25,
		"WallHMax": 0.4,

		"StepSpacingMin": 3,
		"StepSpacingMax": 4,
		"StepH0Min": 0.1,
		"StepH0Max": 0.3,
		"StepH1Min": -0.3,
		"StepH1Max": -0.1,

		"SlopeDeltaRange": 0.05,
		"SlopeDeltaMin": -0.5,
		"SlopeDeltaMax": 0.5
	}
]
}
\end{lstlisting}

\section{Environments}
\label{sec:ModelBasedActionExploration}

Here we describe the types of simulation \environments included in \terrainRLSim.
In total there are almost $100$ environments.

\subsection{TerrainRL}

\begin{figure}[tbh]
	\begin{centering}
	\subcaptionbox{\mixed  \ terrain}{   \includegraphics[width=0.96\columnwidth]{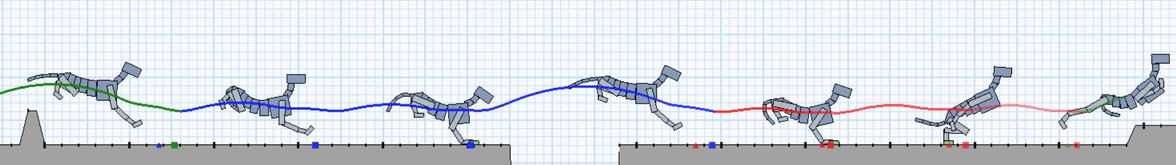}}\\
	\subcaptionbox{\slopesMixed  \ terrain}{   \includegraphics[width=0.96\columnwidth]{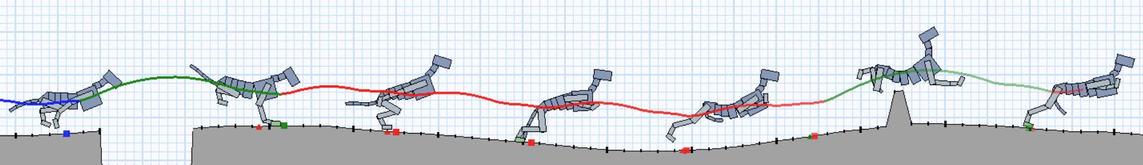}}\\
	\subcaptionbox{\narrowGaps  \ terrain}{   \includegraphics[width=0.96\columnwidth]{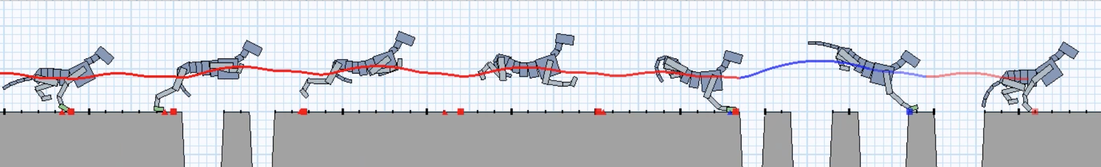}}\\
	\subcaptionbox{\tightGaps  \ terrain}{   \includegraphics[width=0.96\columnwidth]{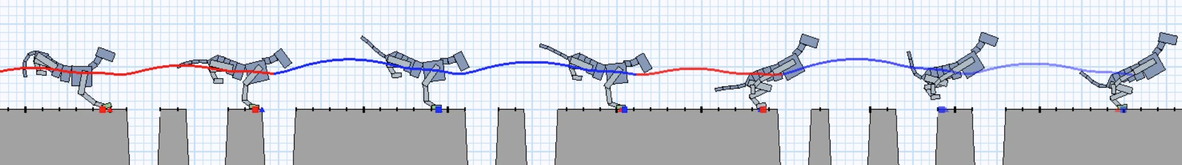}}\\
	\subcaptionbox{\slopesGaps  \ terrain}{   \includegraphics[width=0.96\columnwidth]{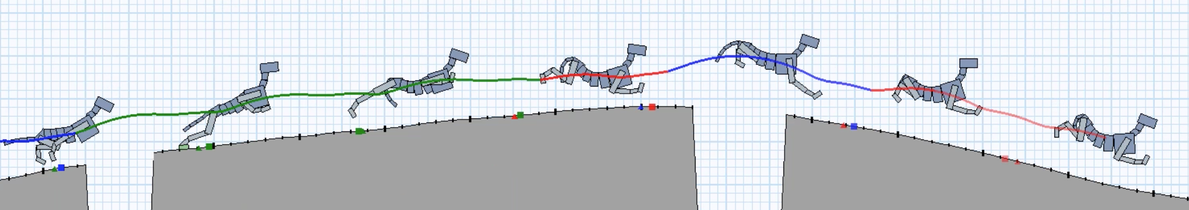}}\\
	\subcaptionbox{\slopesSteps \  terrain}{   \includegraphics[width=0.96\columnwidth]{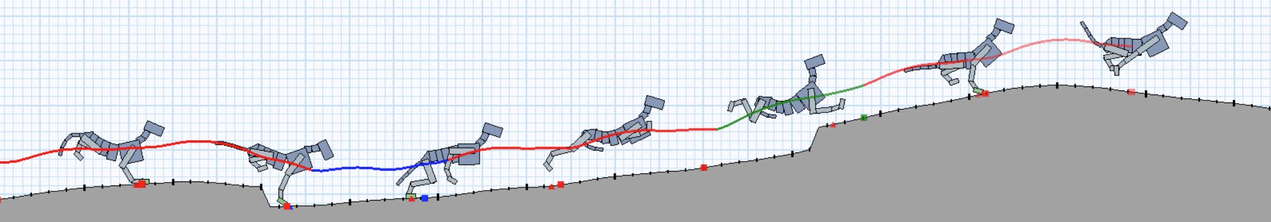}}\\
	\subcaptionbox{\slopesWalls  \ terrain}{   \includegraphics[width=0.96\columnwidth]{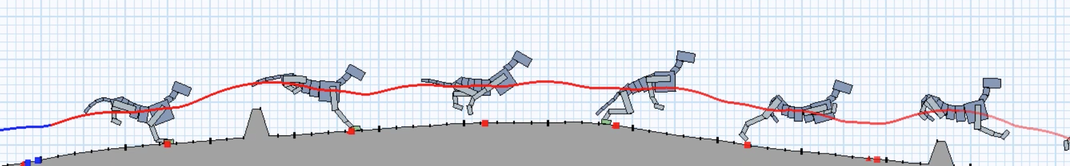}}\\
 \caption{Dog on different terrain types.
   \label{fig:dog-terrains}
 }
	\end{centering}
	\end{figure}

\begin{figure}[tbh]
	\begin{centering}
	\subcaptionbox{Raptor \mixed \ terrain}{   \includegraphics[width=0.96\columnwidth]{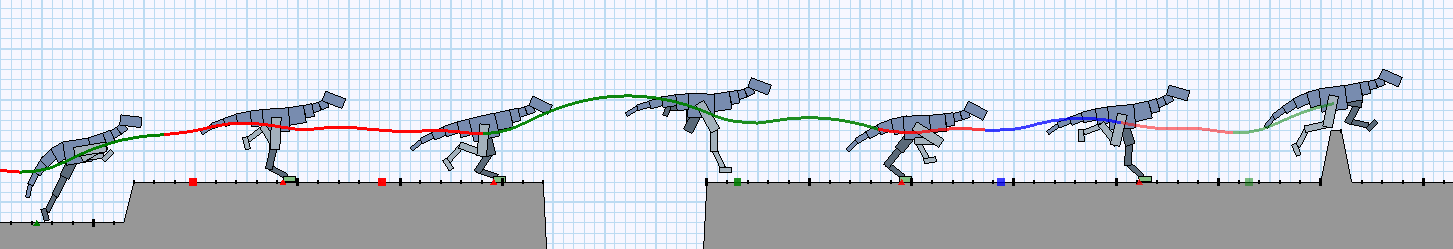}}\\
	\subcaptionbox{Raptor \slopesMixed \ terrain}{   \includegraphics[width=0.96\columnwidth]{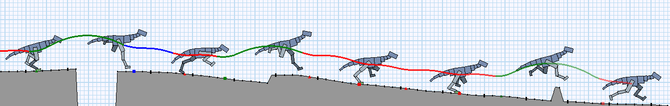}}\\
	\subcaptionbox{Raptor \narrowGaps \ terrain}{   \includegraphics[width=0.96\columnwidth]{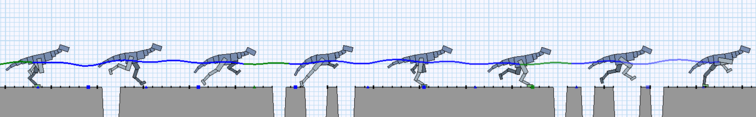}}\\
	\subcaptionbox{Goat on \variableSteps \ terrain}{   \includegraphics[width=0.75\columnwidth]{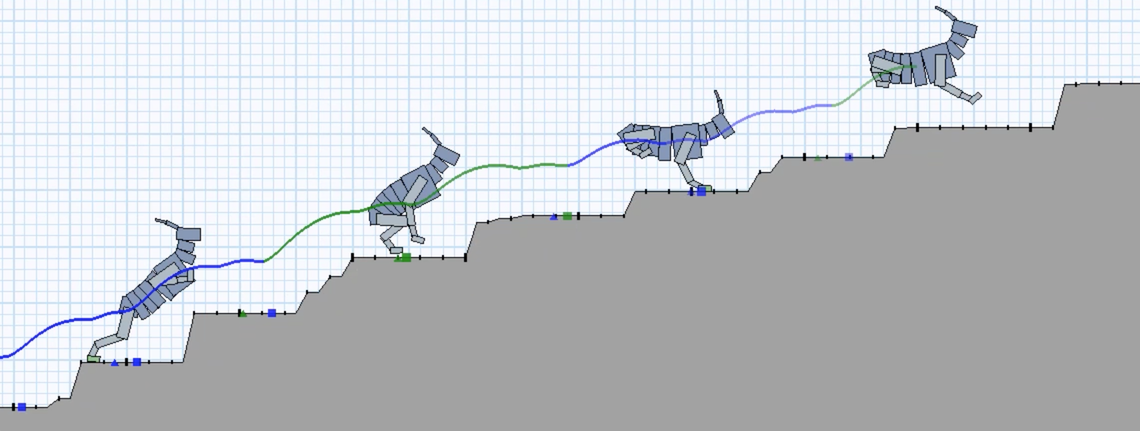}}\\
 \caption{Other Control Policies
   \label{fig:others-on-terrians}
 }
	\end{centering}
	\end{figure}

The \textit{TerrainRL} \environments are based on the work in~\citep{2016-TOG-terrainDeepRL}.
In this work physics-based characters with \FSM controllers are parameterized and trained to traverse complex dynamically generated terrains. 
Examples of different terrain types and characters are shown in~\refFigure{fig:dog-terrains} and~\refFigure{fig:others-on-terrians}.

	On top of what was created for the \terrainRL project we include additional character and terrain types. 
	These include a \SIMBICON based biped controller and a hopper controller. 
	The new terrain types include \textit{cliffs} and other more challenging versions of the ones used in the paper~\citep{2015-TOG-terrainRL}.

\subsection{Imitation Learning}
\label{subsec:imitation-learning}

\begin{figure}[tbh]
	\begin{centering}
	\subcaptionbox{   }{\includegraphics[width=0.2\columnwidth]{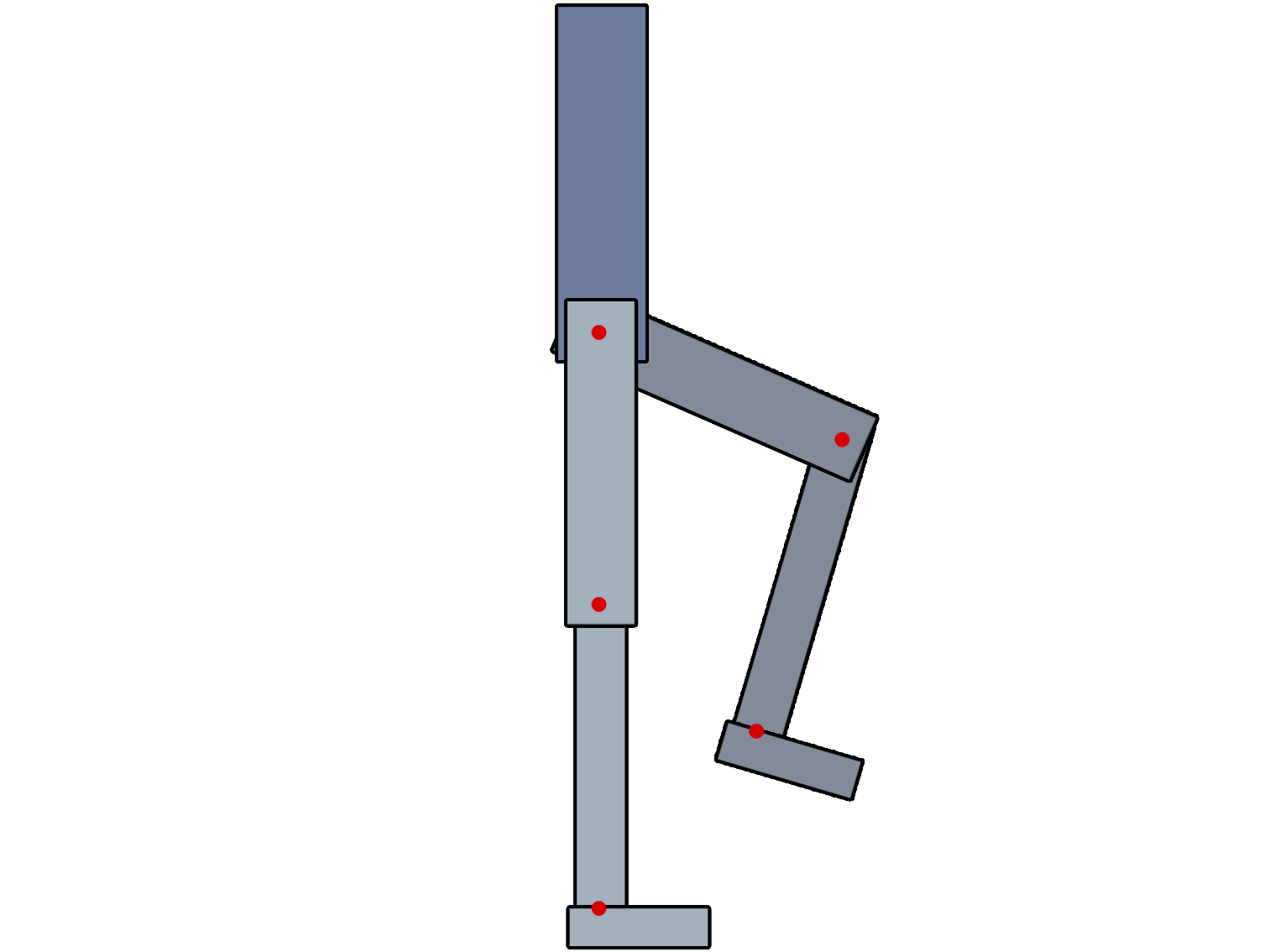}}
	\subcaptionbox{   }{\includegraphics[width=0.25\columnwidth]{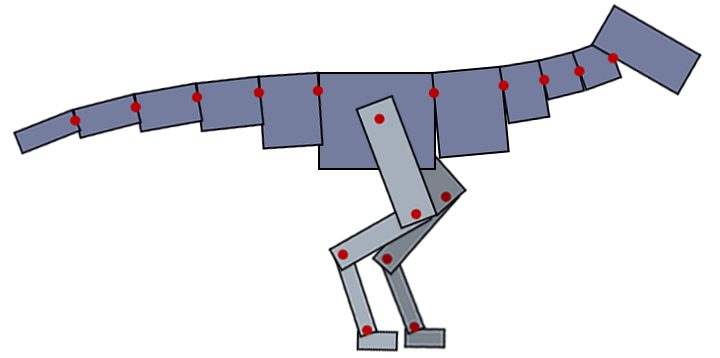}}
	\hspace*{0.25cm} \subcaptionbox{  }{ \includegraphics[width=0.18\columnwidth]{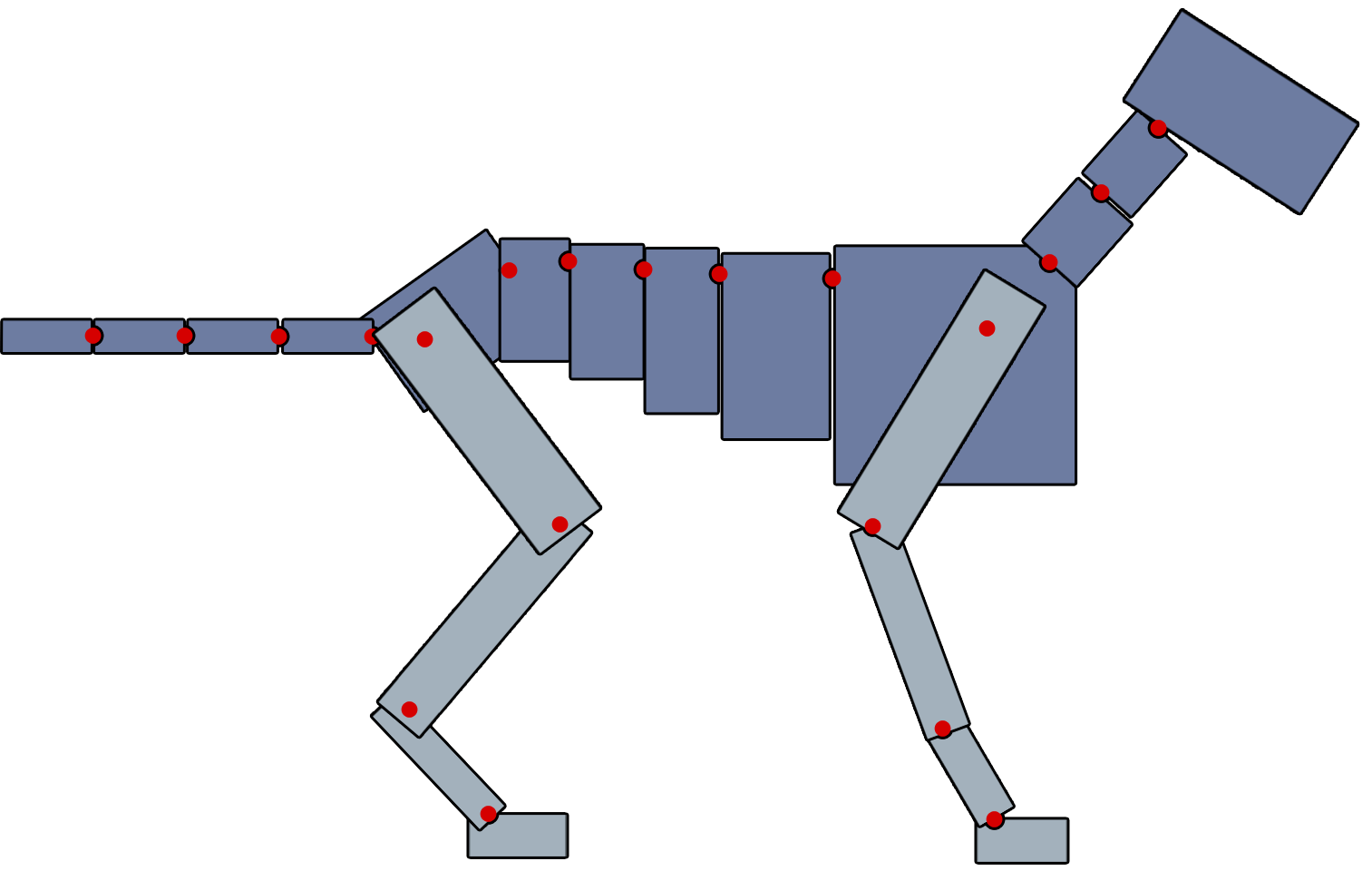}}
	\hspace*{0.4cm} 
	\subcaptionbox{  }{ \includegraphics[width=0.2\columnwidth]{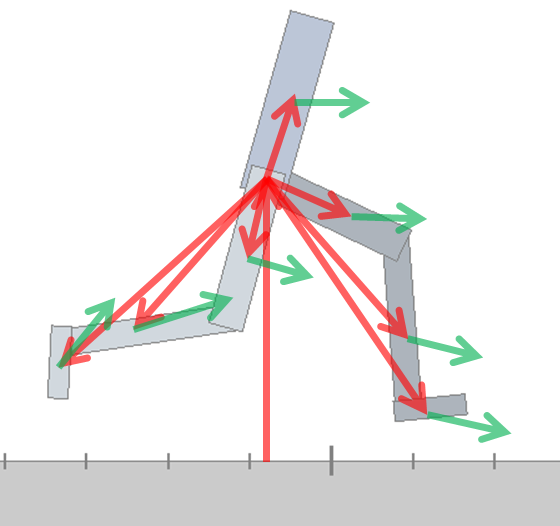}}
 \caption{Simulated articulated figures and their state representation. Revolute joints connect all links. From left to right: 
 7-link \bipedController; 19-link \raptorController; 21-link \dogController; State features: root height, relative position (red) of each link with respect to the root and their respective linear velocity (green).
   \label{fig:characters}   }
	\end{centering}
	\end{figure}

\begin{figure}[h]
	\begin{centering}
	\subcaptionbox{   }{\includegraphics[width=0.65\columnwidth]{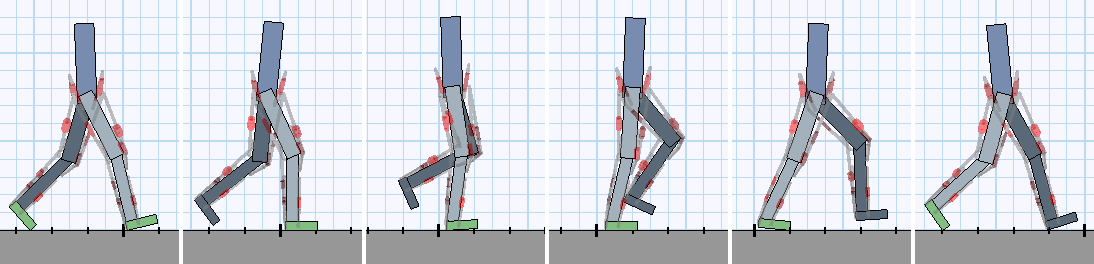}}
	\subcaptionbox{   }{\includegraphics[width=0.65\columnwidth]{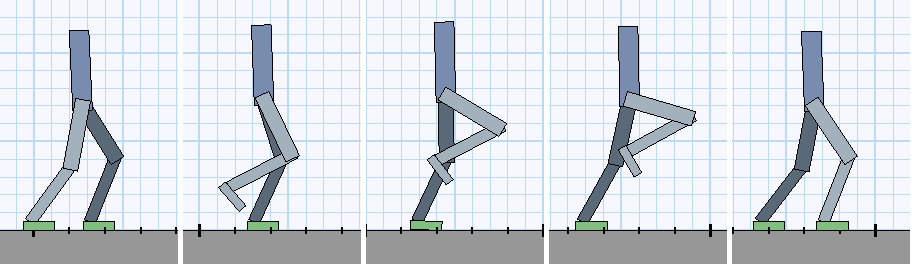}}
	\subcaptionbox{   }{\includegraphics[width=0.65\columnwidth]{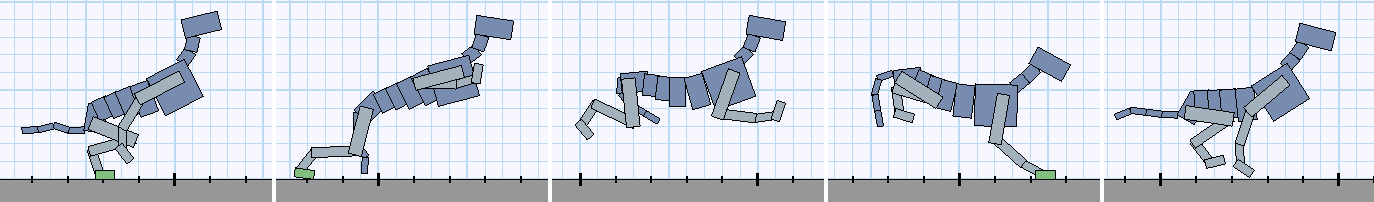}}
	\subcaptionbox{   }{\includegraphics[width=0.65\columnwidth]{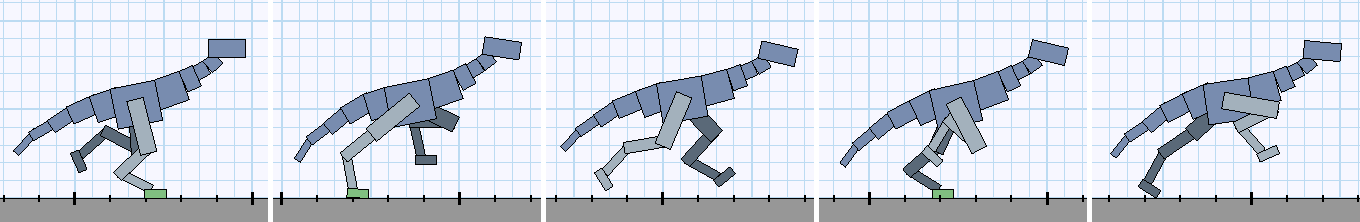}}
 \caption{Simulated Motions Using the desired position Action Representation. The top row uses an muscel action space
  while the remainder are driven by a desired position action space.
        \label{fig:motions}   }
	\end{centering}
	\end{figure}

The goal in these \environments is to train an \agent to imitate particular behaviours described by a motion capture clip, and is based on the work in~\citep{Peng:2017:LLS:3099564.3099567}. 
The provided clip includes sequential character poses that are used in the reward function to instruct the character to match the motion capture pose.
For these environments there are three types of characters that are used, a \bipedController, \raptorController and \dogController (\refFigure{fig:characters}).
For each of these characters there are $4$ different action models available to actuate the joints: torques, desired velocity, desired position and muscle-based control.
Example motions learned on these models are shown in~\refFigure{fig:motions}.

			
	We include additional \environments for learning walking and running motions for 3D bipeds.
	There are also a number of terrain types, including \textit{rough} and \textit{steps}, that can be used to add randomly generated terrain into the simulation.
	
\subsection{DeepLoco}

\begin{figure}[htb]
	\begin{centering}
	\subcaptionbox{\label{fig:deeploco-environemnts-soccer} \soccer }{\includegraphics[width=0.45\columnwidth]{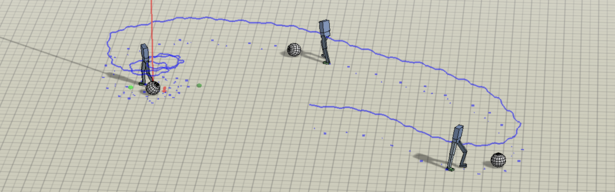}}
	\subcaptionbox{\label{fig:deeploco-environemnts-trail} \pathFollowing}{\includegraphics[width=0.45\columnwidth]{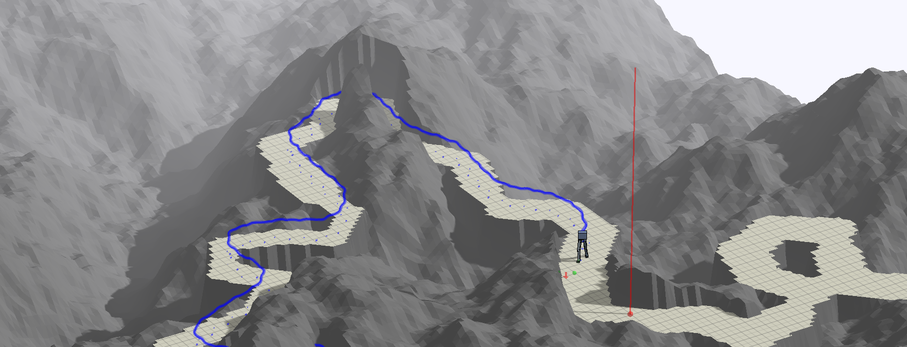}} \\
	\subcaptionbox{\label{fig:deeploco-environemnts-forrest} \forest}{\includegraphics[width=0.45\columnwidth]{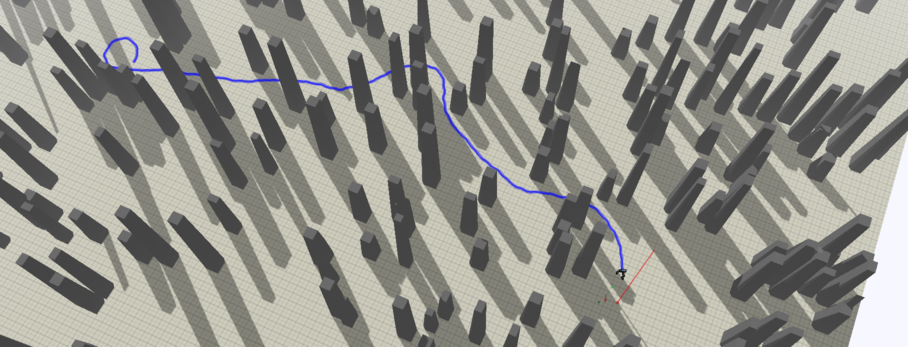}}
	\subcaptionbox{\label{fig:deeploco-environemnts-blocks} \largeBlocks}{\includegraphics[width=0.45\columnwidth]{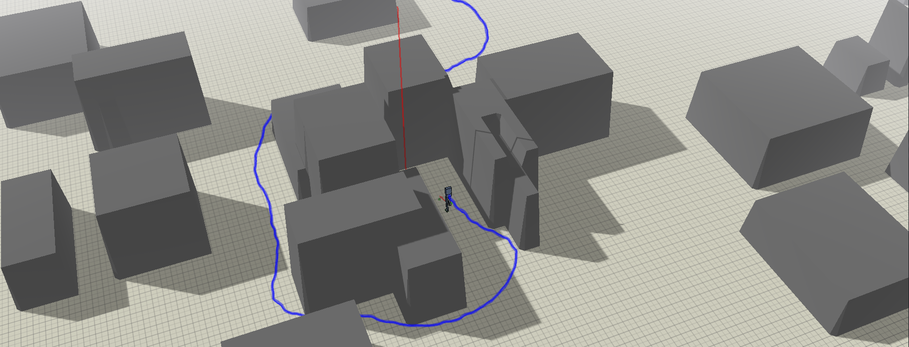}} \\
	\subcaptionbox{\label{fig:deeploco-environemnts-dynamicObs} \dynamicObstacles}{\includegraphics[width=0.45\columnwidth]{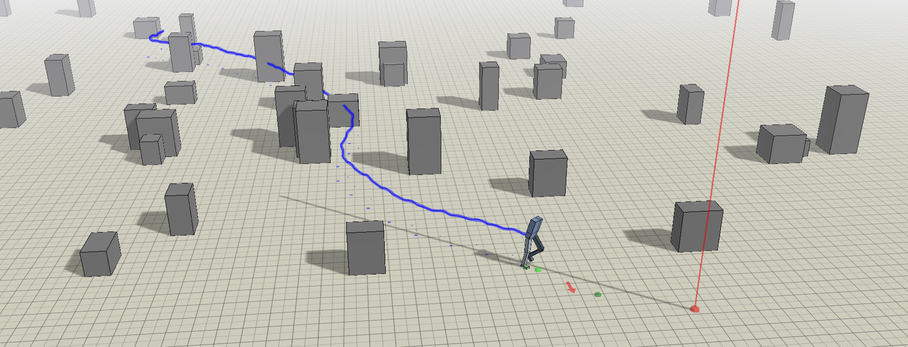}}
	
	\caption{Snapshots of \deepLoco tasks. The red marker represents the target location and the blue line traces the trajectory of the character's centre of mass. \textbf{in order:} soccer dribbling, path following, pillar obstacles, block obstacles, dynamic obstacles.}
	\label{fig:hlcSnapshots}
	\end{centering}
	\end{figure}

The \textit{DeepLoco} \environments are similar to the ones used in~\citep{Peng:2017:DDL:3072959.3073602}.
They include a number of 3D simulations where the goal is to train a biped to walk in complex environments with randomly generated terrain~\refFigure{fig:hlcSnapshots}.

	We include additional environments and configuration that were used for testing and evaluation in the process of completing this project. 
	These include more challenging environments and a version of the controller that does not use hierarchical control, such as a controller that includes the terrain input and operates at \valueWithUnits{30}{fps}. 
	The code also includes processed versions of mocap clips.

\subsection{PLAiD}

\begin{figure}[!ht]
		\centering 
		\subcaptionbox{\label{fig:distilation-environemnts-flat} \Flat}{ \includegraphics[trim={7.5cm 3.5cm 7.5cm 3.5cm},clip,width=0.32\columnwidth]{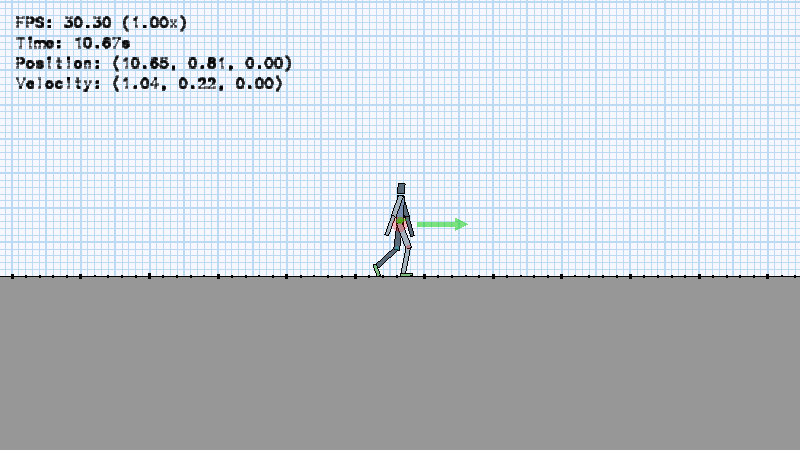}}
		\subcaptionbox{\label{fig:distilation-environemnts-incline} \incline}{ \includegraphics[trim={7.5cm 3.5cm 7.5cm 3.5cm},clip,width=0.32\columnwidth]{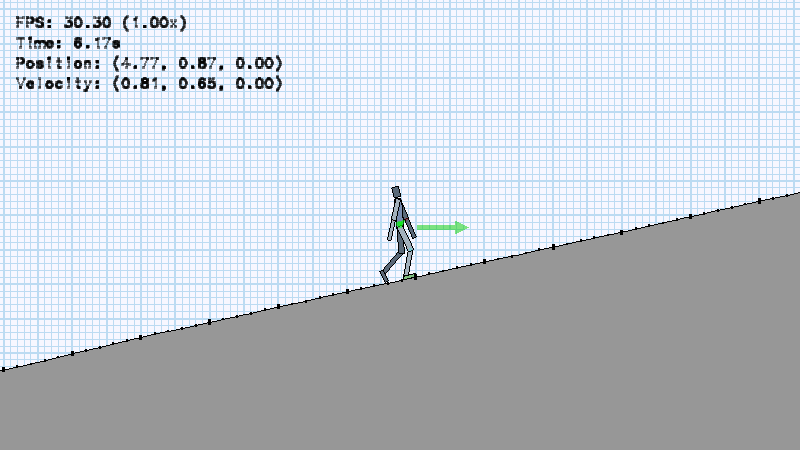}}
		\subcaptionbox{\label{fig:distilation-environemnts-steps} \steps}{ \includegraphics[trim={7.5cm 3.5cm 7.5cm 3.5cm},clip,width=0.32\columnwidth]{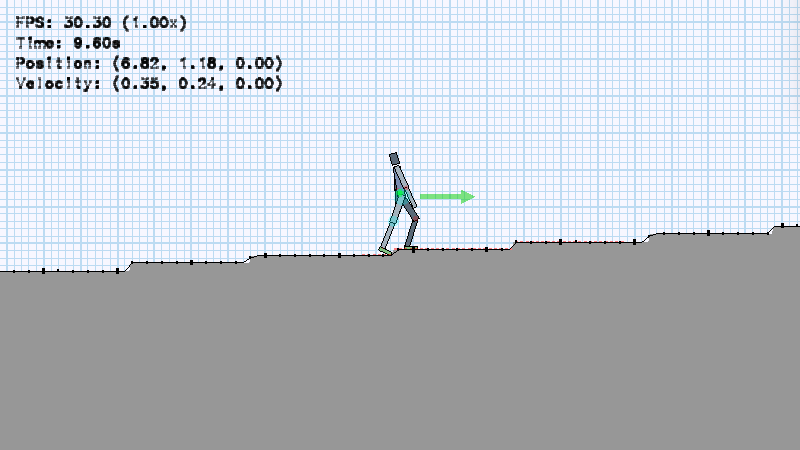}} \\
		\subcaptionbox{\label{fig:distilation-environemnts-slopes} \slopes}{ \includegraphics[trim={2.5cm 0.0cm 2.5cm 3.5cm},clip,width=0.32\columnwidth]{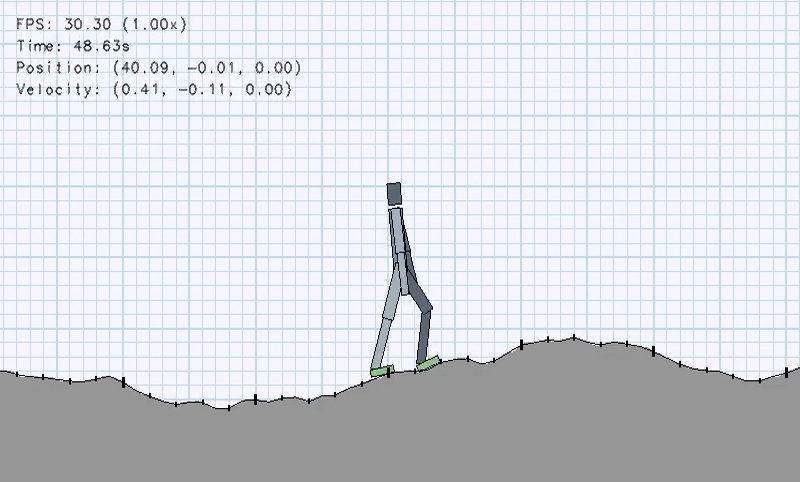}}
		\subcaptionbox{\label{fig:distilation-environemnts-gaps} \gaps}{ \includegraphics[trim={2.5cm 0.0cm 2.5cm 3.5cm},clip,width=0.32\columnwidth]{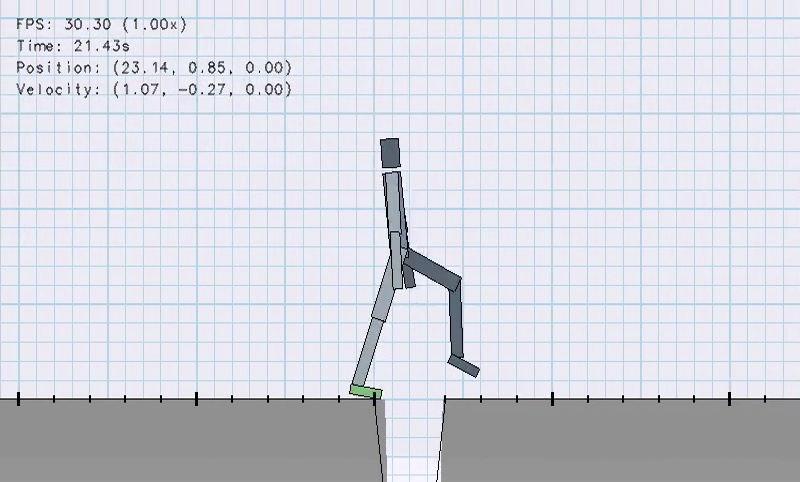}} 
		\subcaptionbox{\label{fig:distilation-environemnts-mixed} \mixed}{ \includegraphics[trim={2.5cm 0.0cm 2.5cm 3.5cm},clip,width=0.32\columnwidth]{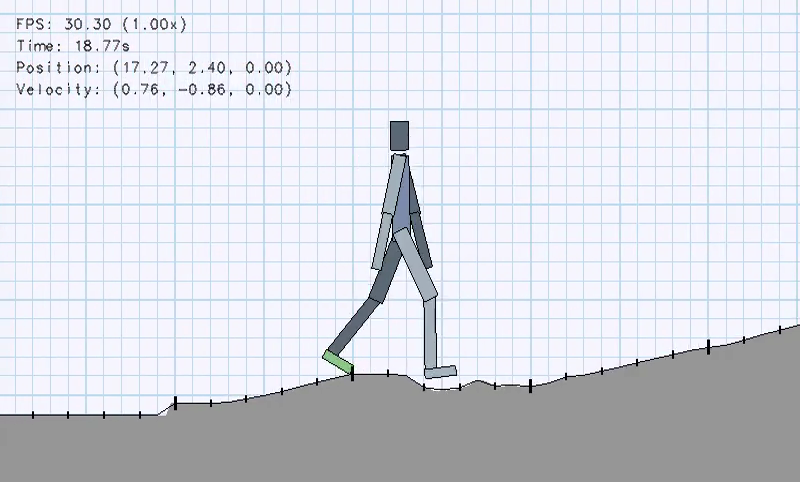}} \\
		\caption{
		The environments used to evaluate \progRL.
		}
		\label{fig:distilation-environemnts}
	\end{figure}

These environments are an extension of the \environments in~\refSection{subsec:imitation-learning}. 
Here the \agent has been modified to have arms and the terrain is randomly generated.
With the addition of randomly generated terrain additional state features are added to provide visual perception of the terrain.
Part of these environments were used in~\citep{2018arXiv180204765B}.
These are the only available environments that can be used for multi-task and continual learning in the continuous action space domain for \RL.
Examples of the \environments are shown in~\refFigure{fig:distilation-environemnts}.
	
	On top of the environments used in \progRL we add additional ones for both the \dogController and \raptorController characters.
	These additional \environments are for each of the terrain types in~\refFigure{fig:distilation-environemnts} and two additional terrain types, \textit{walls} and \textit{slopes-mixed}. 
	New \environments were also created for a 3D biped with different 3D terrain types.
	These \environments contain most of the challenging aspects of interesting problems and are well suited for the use of comparing \deepRL methods.

\section{Discussion}
\label{sec:discussion}

Many of these \environments have been used to create robust controllers that produce high quality motion. 
Even with the great progress in this area there is still much work to be done.
These \environments use more realistic joint torque limits that a true biological version of the character might be capable of.
Having realistic torque limits is a start but only captures a small portion of the complexities of generating torques.
For biological creatures the maximal and minimal possible joint torques can be different depending on the direction of rotation, it can even depend on the joint pose.
In many robotics applications you have to cope with the issue of \textit{backlash} that involves the amount of free space between the engineered parts causing the system to move in unintended directions.
There is also the complexity of flex in the system which is sometimes intended as springs are used to absorb forces.
It is possible to model many of these phenomenon in a physics simulation already.

Apart from the physical phenomenon that we can and should modelled better in simulations used for \RL there appear to be a number of simulation parameters that could be given values more in-line with the real world.
Examples of these include: linear dampening, gravity, angular dampening, static and kinetic friction values, proper masses and densities of objects, etc.
Often, the more the simulation is constructed to model the real world the more challenging solving it becomes.
It would increase the community benefit to evaluate \RL methods on environments that have more purpose and can be used in games an on robots.

As we further \RL research we should also be pushing the simulation accuracy and task difficulty to help us converge on solutions that will work in the real world.
There is a great deal of work left to be done before we can create human level intelligence on both more accurate simulation models and better learning techniques.

\todo{Should train using plaid slopes-mixed}

\bibliographystyle{tex/named}
\bibliography{paper}

\begin{thebibliography}{}

\bibitem[\protect\citeauthoryear{Beattie \bgroup \em et al.\egroup
  }{2016}]{DBLP:journals/corr/BeattieLTWWKLGV16}
Charles Beattie, Joel~Z. Leibo, Denis Teplyashin, Tom Ward, Marcus Wainwright,
  Heinrich K{\"{u}}ttler, Andrew Lefrancq, Simon Green, V{\'{\i}}ctor
  Vald{\'{e}}s, Amir Sadik, Julian Schrittwieser, Keith Anderson, Sarah York,
  Max Cant, Adam Cain, Adrian Bolton, Stephen Gaffney, Helen King, Demis
  Hassabis, Shane Legg, and Stig Petersen.
\newblock Deepmind lab.
\newblock {\em CoRR}, abs/1612.03801, 2016.

\bibitem[\protect\citeauthoryear{Bellemare \bgroup \em et al.\egroup
  }{2012}]{DBLP:journals/corr/abs-1207-4708}
Marc~G. Bellemare, Yavar Naddaf, Joel Veness, and Michael Bowling.
\newblock The arcade learning environment: An evaluation platform for general
  agents.
\newblock {\em CoRR}, abs/1207.4708, 2012.

\bibitem[\protect\citeauthoryear{{Berseth} \bgroup \em et al.\egroup
  }{2018}]{2018arXiv180204765B}
G.~{Berseth}, C.~{Xie}, P.~{Cernek}, and M.~{Van de Panne}.
\newblock {Progressive Reinforcement Learning with Distillation for
  Multi-Skilled Motion Control}.
\newblock {\em ArXiv e-prints}, February 2018.

\bibitem[\protect\citeauthoryear{Brockman \bgroup \em et al.\egroup
  }{2016}]{DBLP:journals/corr/BrockmanCPSSTZ16}
Greg Brockman, Vicki Cheung, Ludwig Pettersson, Jonas Schneider, John Schulman,
  Jie Tang, and Wojciech Zaremba.
\newblock Openai gym.
\newblock {\em CoRR}, abs/1606.01540, 2016.

\bibitem[\protect\citeauthoryear{Bullet}{2015}]{Bullet}
Bullet.
\newblock Bullet physics library, 2015.
\newblock http://bulletphysics.org.

\bibitem[\protect\citeauthoryear{Gibson}{1979}]{gibson1979ecological}
James~Jerome Gibson.
\newblock The ecological approach to visual perception.
\newblock 1979.

\bibitem[\protect\citeauthoryear{Kajita \bgroup \em et al.\egroup
  }{2001}]{973365}
S.~Kajita, F.~Kanehiro, K.~Kaneko, K.~Yokoi, and H.~Hirukawa.
\newblock The 3d linear inverted pendulum mode: a simple modeling for a biped
  walking pattern generation.
\newblock In {\em Proceedings 2001 IEEE/RSJ International Conference on
  Intelligent Robots and Systems. Expanding the Societal Role of Robotics in
  the the Next Millennium (Cat. No.01CH37180)}, volume~1, pages 239--246 vol.1,
  2001.

\bibitem[\protect\citeauthoryear{Kajita \bgroup \em et al.\egroup
  }{2003}]{1241826}
S.~Kajita, F.~Kanehiro, K.~Kaneko, K.~Fujiwara, K.~Harada, K.~Yokoi, and
  H.~Hirukawa.
\newblock Biped walking pattern generation by using preview control of
  zero-moment point.
\newblock In {\em 2003 IEEE International Conference on Robotics and Automation
  (Cat. No.03CH37422)}, volume~2, pages 1620--1626 vol.2, Sept 2003.

\bibitem[\protect\citeauthoryear{{Mania} \bgroup \em et al.\egroup
  }{2018}]{2018arXiv180307055M}
H.~{Mania}, A.~{Guy}, and B.~{Recht}.
\newblock {Simple random search provides a competitive approach to
  reinforcement learning}.
\newblock {\em ArXiv e-prints}, March 2018.

\bibitem[\protect\citeauthoryear{Peng and van~de
  Panne}{2017}]{Peng:2017:LLS:3099564.3099567}
Xue~Bin Peng and Michiel van~de Panne.
\newblock Learning locomotion skills using deeprl: Does the choice of action
  space matter?
\newblock In {\em Proceedings of the ACM SIGGRAPH / Eurographics Symposium on
  Computer Animation}, SCA '17, pages 12:1--12:13, New York, NY, USA, 2017.
  ACM.

\bibitem[\protect\citeauthoryear{Peng \bgroup \em et al.\egroup
  }{2015}]{2015-TOG-terrainRL}
Xue~Bin Peng, Glen Berseth, and Michiel van~de Panne.
\newblock Dynamic terrain traversal skills using reinforcement learning.
\newblock {\em ACM Transactions on Graphics}, 34(4):Article 80, 2015.

\bibitem[\protect\citeauthoryear{Peng \bgroup \em et al.\egroup
  }{2016}]{2016-TOG-terrainDeepRL}
Xue~Bin Peng, Glen Berseth, and Michiel van~de Panne.
\newblock Terrain-adaptive locomotion skills using deep reinforcement learning.
\newblock {\em ACM Transactions on Graphics}, 35(4):Article 81, 2016.

\bibitem[\protect\citeauthoryear{Peng \bgroup \em et al.\egroup
  }{2017}]{Peng:2017:DDL:3072959.3073602}
Xue~Bin Peng, Glen Berseth, Kangkang Yin, and Michiel Van De~Panne.
\newblock Deeploco: Dynamic locomotion skills using hierarchical deep
  reinforcement learning.
\newblock {\em ACM Trans. Graph.}, 36(4):41:1--41:13, July 2017.

\bibitem[\protect\citeauthoryear{Rajeswaran \bgroup \em et al.\egroup
  }{2017}]{DBLP:journals/corr/RajeswaranLTK17}
Aravind Rajeswaran, Kendall Lowrey, Emanuel Todorov, and Sham Kakade.
\newblock Towards generalization and simplicity in continuous control.
\newblock {\em CoRR}, abs/1703.02660, 2017.

\bibitem[\protect\citeauthoryear{{Salimans} \bgroup \em et al.\egroup
  }{2017}]{2017arXiv170303864S}
T.~{Salimans}, J.~{Ho}, X.~{Chen}, S.~{Sidor}, and I.~{Sutskever}.
\newblock {Evolution Strategies as a Scalable Alternative to Reinforcement
  Learning}.
\newblock {\em ArXiv e-prints}, March 2017.

\bibitem[\protect\citeauthoryear{{Tassa} \bgroup \em et al.\egroup
  }{2018}]{2018arXiv180100690T}
Y.~{Tassa}, Y.~{Doron}, A.~{Muldal}, T.~{Erez}, Y.~{Li}, D.~{de Las Casas},
  D.~{Budden}, A.~{Abdolmaleki}, J.~{Merel}, A.~{Lefrancq}, T.~{Lillicrap}, and
  M.~{Riedmiller}.
\newblock {DeepMind Control Suite}.
\newblock {\em ArXiv e-prints}, January 2018.

\bibitem[\protect\citeauthoryear{Tian \bgroup \em et al.\egroup
  }{2017}]{NIPS2017_6859}
Yuandong Tian, Qucheng Gong, Wenling Shang, Yuxin Wu, and C.~Lawrence Zitnick.
\newblock Elf: An extensive, lightweight and flexible research platform for
  real-time strategy games.
\newblock In I.~Guyon, U.~V. Luxburg, S.~Bengio, H.~Wallach, R.~Fergus,
  S.~Vishwanathan, and R.~Garnett, editors, {\em Advances in Neural Information
  Processing Systems 30}, pages 2659--2669. Curran Associates, Inc., 2017.

\bibitem[\protect\citeauthoryear{Todorov \bgroup \em et al.\egroup
  }{2012}]{6386109}
E.~Todorov, T.~Erez, and Y.~Tassa.
\newblock Mujoco: A physics engine for model-based control.
\newblock In {\em 2012 IEEE/RSJ International Conference on Intelligent Robots
  and Systems}, pages 5026--5033, Oct 2012.

\bibitem[\protect\citeauthoryear{Yamaguchi \bgroup \em et al.\egroup
  }{1999}]{770006}
J.~Yamaguchi, E.~Soga, S.~Inoue, and A.~Takanishi.
\newblock Development of a bipedal humanoid robot-control method of whole body
  cooperative dynamic biped walking.
\newblock In {\em Proceedings 1999 IEEE International Conference on Robotics
  and Automation (Cat. No.99CH36288C)}, volume~1, pages 368--374 vol.1, 1999.

\bibitem[\protect\citeauthoryear{Yin \bgroup \em et al.\egroup }{2007}]{Yin07}
KangKang Yin, Kevin Loken, and Michiel van~de Panne.
\newblock Simbicon: Simple biped locomotion control.
\newblock {\em ACM Transctions on Graphics}, 26(3):Article 105, 2007.

\end{thebibliography}

\end{document}